# Deep Learning 2.0: Artificial Neurons That Matter - Reject Correlation, Embrace Orthogonality

Taha Bouhsine
MLNomads
Agadir, Morocco
`yat@mlnomads.com`

## Abstract

*We introduce $\mathbb{E}$-product-powered neural network, the Neural Matter Network (NMN), a breakthrough in deep learning that achieves non-linear pattern recognition without activation functions. Our key innovation relies on the $\mathbb{E}$-product and $\overline{\mathbb{E}}$-product, which naturally induces non-linearity by projecting inputs into a pseudo-metric space, eliminating the need for traditional activation functions while maintaining only a softmax layer for final class probability distribution. This approach simplifies network architecture and provides unprecedented transparency into the network's decision-making process. Our comprehensive empirical evaluation across different datasets demonstrates that NMN consistently outperforms traditional MLPs. The results challenge the assumption that separate activation functions are necessary for effective deep-learning models. The implications of this work extend beyond immediate architectural benefits: by eliminating intermediate activation functions while preserving non-linear capabilities, $\mathbb{E}$-MLP establishes a new paradigm for neural network design that combines simplicity with effectiveness. Most importantly, our approach provides unprecedented insights into the traditionally opaque "black-box" nature of neural networks, offering a clearer understanding of how these models process and classify information.*

## 1. Introduction

The perceptron, introduced by Rosenblatt in 1958 [26], has played a fundamental role in the development of neural networks, serving as an essential building block in artificial intelligence over the past six decades [11, 24]. As a linear model, the perceptron projects input data onto an Euclidean space where the dot product of inputs and weights capture similarity. However, the perceptron's linear nature limits its ability to model complex, non-linear patterns in data [19, 29].

The incorporation of non-linear activation functions, such as the Rectified Linear Unit (ReLU) [21], was a pivotal advancement, enabling neural networks to approximate a wide range of functions and thus address the limitations inherent in linear models [10]. This transformation has allowed for the development of deep learning architectures that can capture intricate relationships within large datasets. However, this gain in representational flexibility has introduced a significant trade-off: as models grow in complexity, the interpretability of their learned representations diminishes, making it increasingly challenging to understand the underlying decision-making processes [14, 16, 27].

A core challenge lies in the transformation of data space. The dot product in the perceptron layer, rooted in Euclidean geometry, measures the similarity between inputs and weight vectors. When followed by non-linear activation functions, this transformation projects data into spaces that are no longer well-defined or explainable, obscuring insights into the learned representations. As neural network models grow in complexity, they become less interpretable, making it challenging to fully comprehend how decisions are made, especially in high-stakes applications [5].

In this paper, we introduce the $\mathbb{E}$-product as a novel solution to these challenges, aiming to bridge the gap between model complexity and interpretability. The $\mathbb{E}$-product combines squared Euclidean distance with orthogonality through the squared dot product. Originally proposed by Bouhsine et al. in the context of contrastive learning [1], this operation works in a pseudo-metric space [2] which retains non-linearity without relying on activation functions.

Building on this innovation, we propose the Neural-Matter Network (NMN), a new network layer that leverages the $\mathbb{E}$-product to create deep neural networks without activation functions. NMNs operate in a pseudo-metric space that accommodates intricate, in-terdependent data relationships without distorting the geometric topology of the data. By eliminating the need for activation functions, NMNs simplify network architecture, making it possible to interpret weight dependencies directly without sacrificing the model's capacity to learn complex patterns.

Our work also introduces the Neural-Matter State (NMS) Plots, a new framework for visualizing and interpreting weight distributions within NMNs. For the first time, this approach attempts to provide a framework to explore the "black box" of neural networks, offering insights into the organization and significance of learned weights. The NMS Plot reveals not only how individual weights contribute to model predictions but also how weight clusters influence factors like overfitting and feature uniqueness. This advance holds significant implications for enhancing the interpretability and trustworthiness of AI systems as they are increasingly applied to critical decision-making tasks.

Additionally, we introduce the Ǝ-regularizer, which is inspired by the SimO loss proposed by Bouhsine et al. [1]. This regularizer addresses the issue of neural collapse, where neurons in the model become too close in the neuron space. By promoting orthogonality among neurons, the Ǝ-regularizer optimizes their spatial to ensure that they are positioned further apart from each other and non-linearly dependent then each other.

Our contributions are as follows:

- The Neural-Matter Network (NMN), is an activation-free network architecture using a non-linear Ǝ-product to learn representations without distorting data topology.
- Introducing Ǝ-regularization for intra-orthogonality vectors in the weight matrix.
- Introduction of the Neural-Matter State (NMS) Plot, a framework for interpreting and visualizing learned weights.
- Affero GNU Open-source implementation of the Neural-Matter Layer and related experiments using Flax linen/Jax.

These contributions mark a significant advancement in creating more interpretable and potentially efficient deep learning models. By introducing Ǝ-product-based architectures, we pave the way for further exploration of activation-free designs and more explainable deep learning frameworks.

## 2. Theoretical Foundation

### 2.1. Multi-Layer Perceptron: Theoretical Analysis

#### 2.1.1 Fundamental Components

Consider an input vector $\mathbf{x} \in \mathbb{R}^n$ entering an MLP layer. Each neuron is characterized by a weight vector $\mathbf{w} \in \mathbb{R}^n$ and a bias term $b \in \mathbb{R}$. The transformation process consists of two primary operations:

1. **Affine Transformation:** The neuron computes a weighted sum followed by a bias addition:

$$z = \mathbf{w}^T \mathbf{x} + b = \sum_{i=1}^{n} w_i x_i + b \qquad (1)$$

2. **Non-linear Activation:** The affine result undergoes non-linear transformation via function $f : \mathbb{R} \to \mathbb{R}$:

$$y = f(z) = f(\mathbf{w}^T \mathbf{x} + b) \qquad (2)$$

For a layer containing $m$ neurons, we express the operation in matrix notation. Let $\mathbf{W} \in \mathbb{R}^{m \times n}$ denote the weight matrix where each row corresponds to a neuron's weight vector, and let $\mathbf{b} \in \mathbb{R}^m$ represent the bias vector. For a batch of $k$ input samples $\mathbf{X} \in \mathbb{R}^{k \times n}$, the layer output $\mathbf{Y} \in \mathbb{R}^{k \times m}$ is computed as:

$$\mathbf{Y} = f(\mathbf{X}\mathbf{W}^T + \mathbf{b}) \qquad (3)$$

In traditional MLP architectures, vector similarity is computed between the input vector $\mathbf{x}$ and each weight vector $\mathbf{w}_i$ using the dot product. For neuron $i$:

$$z_i = \mathbf{x} \cdot \mathbf{w}_i = \sum_{j=1}^{n} x_j w_{ij} \qquad (4)$$

For illustration, consider $\mathbf{x} = [1, 2, 3]$ and $\mathbf{w}_i = [0.5, -1, 0.2]$:

$$z_i = (1 \times 0.5) + (2 \times -1) + (3 \times 0.2) = -0.9 \qquad (5)$$

The activation function introduces non-linearity into the network. Using ReLU as an example:

$$a_i = \max(0, z_i) \qquad (6)$$

For instance, given $\mathbf{z} = [-0.9, 1.5, -0.3, 2.1]$, ReLU yields:

$$\mathbf{a} = [0, 1.5, 0, 2.1] \qquad (7)$$

The final layer typically employs softmax to generate probability distributions:

$$\text{softmax}(x_i) = \frac{\exp(x_i)}{\sum_j \exp(x_j)} \qquad (8)$$

For logits $\mathbf{z} = [2.0, 1.0, 0.1]$, softmax produces:

$$\text{softmax}(\mathbf{z}) = [0.665, 0.245, 0.090] \qquad (9)$$

### 2.1.2 Architectural Limitations

**Linear Similarity Constraints** The dot product's linearity in Euclidean space limits expressiveness. While ReLU introduces non-linearity, information loss occurs through negative value elimination.

**Activation Range Issues**: Unbounded dot product outputs ($-\infty$ to $+\infty$) can lead to neuron dominance. For example, when $a \cdot w_1 = 100$ versus $a \cdot w_2 = 0.1$, ReLU preserves magnitude differences, potentially overshadowing subtle patterns.

**Topological Structure Preservation**: The interleaving of linear and non-linear operations can obscure topological relationships within the embedding space, complicating interpretability and structural information preservation.

Dropout serves as a primary mitigation strategy by randomly zeroing activations:

$$\mathbf{a}_{\text{dropout}} = \mathbf{m} \odot \mathbf{a} \tag{10}$$

where $\mathbf{m}$ is a binary mask and $\odot$ denotes element-wise multiplication. For example:

$$\mathbf{a}_{\text{original}} = [0, 1.5, 0, 122.1] \tag{11}$$
$$\mathbf{a}_{\text{dropout}} = [0, 1.5, 0, 0] \tag{12}$$

However, dropout remains a probabilistic solution that does not address fundamental dot product limitations, especially during inference.

These limitations motivate the exploration of alternative similarity measures operating in non-euclidean spaces, capturing both similarity and orthogonality in unified operations.

### 2.2. Bouhsine's Products ($\mathbb{E}$-product and $\overline{\mathbb{E}}$-product)

#### 2.2.1 Definition

The Bouhsine products [1] introduce two distinct similarity measures between vectors in $\mathbb{R}^n$: the $\mathbb{E}$-product and the $\overline{\mathbb{E}}$-product, defined for two vectors $\mathbf{e}_1, \mathbf{e}_2 \in \mathbb{R}^n$ as follows:

**$\mathbb{E}$-product (yat)**

$$\mathbf{e}_1 \; \mathbb{E} \; \mathbf{e}_2 = \frac{(\mathbf{e}_1 \cdot \mathbf{e}_2)^2}{||\mathbf{e}_2 - \mathbf{e}_1||^2} \tag{13}$$

**$\overline{\mathbb{E}}$-product (posi-yat)**

$$\mathbf{e}_1 \; \overline{\mathbb{E}} \; \mathbf{e}_2 = \frac{||\mathbf{e}_2 - \mathbf{e}_1||^2}{(\mathbf{e}_1 \cdot \mathbf{e}_2)^2} \tag{14}$$

Here, $\cdot$ denotes the standard dot product, $||\cdot||$ represents the Euclidean norm.

The Bouhsine products are pseudo-metric (for $\mathbb{E}$) and semi-metric ($\overline{\mathbb{E}}$) (Theorem 8.1), satisfying closure and commutativity, though lacking other conventional algebraic properties such as associativity, distributivity, and the existence of an identity element. This unique structure enables the Bouhsine products to capture aspects of vector relationships that standard similarity metrics like the dot product may overlook.

#### 2.2.2 Limitations of the Dot Product and Advantages of the $\mathbb{E}$-Product

Traditional similarity measures, such as the dot product, often fall short in capturing comprehensive vector relationships, especially when vectors share the same direction but vary in magnitude. This limitation becomes particularly apparent when interpreting neuron weights geometrically, where each weight vector represents a specific reference in the embedding space.

Consider a set of neuron weight vectors $(1, 1)$, $(2, 2)$, $(3, 3)$, $(4, 4)$, $(5, 5)$, $(8, 8)$, and $(9, 9)$, which are all parallel and thus point in the same direction. For a new point $(6, 6)$, using cosine similarity (or dot product) would yield identical results for all these vectors since cosine similarity emphasizes direction while ignoring magnitude. This outcome is misleading, as intuitively, the point $(6, 6)$ is closest to $(5, 5)$ in both direction and magnitude. By disregarding magnitude, the dot product and cosine similarity fail to differentiate the proximity between $(6, 6)$ and the individual weight vectors.

The $\mathbb{E}$-product addresses this limitation by incorporating both magnitude and distance information. Unlike the dot product, which is primarily based on angular similarity, the $\mathbb{E}$-product is designed to account for both directional alignment and the relative distances between vectors. When applying the $\mathbb{E}$-product to compare $(6, 6)$ with each neuron vector, it correctly identifies $(5, 5)$ as the closest match, offering a nuanced understanding that aligns with the intuitive notion of similarity.

Figure 6 visually demonstrates the differences between the dot product and the $\mathbb{E}$-product. Plot (b) shows the dot product's tendency to favor larger vector magnitudes. In contrast, the $\mathbb{E}$-product plot (c) effectively differentiates between vectors based on both magnitude and spatial proximity, underscoring its advantage in applications that require a holistic similarity measure.

### 3. Methods

We propose a novel approach to the Multilayer Perceptron (MLP) layer operation, replacing the traditional dot product with a custom product we call the $\mathbb{E}$-product.

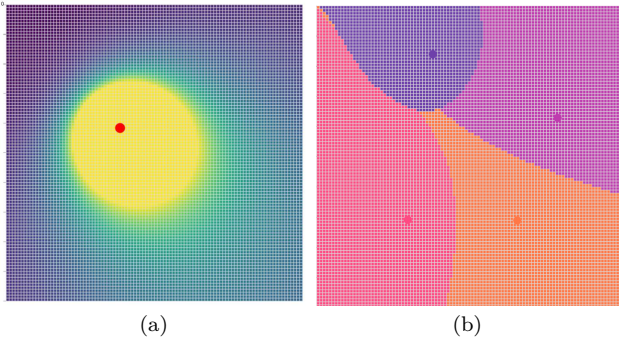

Figure 1. Neuron Field Plot of a (a) single 𝔼-Neuron without scale and (b) multiple 𝔼-Neurons impact on the space

## 3.1. 𝔼-Neuron

In our proposed neuron, we replace the traditional dot product in the MLP layer with the 𝔼-product. For a single 𝔼-neuron (Code 8.3) with weight vector $\mathbf{w} \in \mathbb{R}^n$ and input vector $\mathbf{x} \in \mathbb{R}^n$, the output $y$ is computed as:

$$y = \odot * \mathbb{E}(\mathbf{w}, \mathbf{x}) + b$$

with the $\odot \in \mathbb{R}$ is the scale factor that is equal to:

$$\odot = \left(\frac{n}{log(1+n)}\right)^\alpha$$

The full output of the neuron is:

$$y = \left(\frac{n}{log(1+n)}\right)^\alpha * \frac{(\mathbf{w} \cdot \mathbf{x})^2}{\epsilon + ||\mathbf{x} - \mathbf{w}||^2} + b$$

where $b \in \mathbb{R}$ is the bias term and $\alpha \in \mathbb{R}$ is a learnable parameter for damening of the output and $\epsilon$ is a small positive constant added to ensure numerical stability.

Figure 1 shows the impact of neurons in the embeddings space, instead of learning a boundary, the neuron learns a position in the space that maximizes the 𝔼-score between it and the features vector it is trying to attract.

## 3.2. Neural-Matter State (NMS) Plot

To assess neuron distribution within a layer, we use t-SNE and PCA for dimensionality reduction on neuron weight vectors. This visualization involves a scatter plot with a corresponding 2D density plot and is accompanied by a similarity matrix based on prior work by Bouhsine et al. [1]. Through these representations, we analyze neuron alignment and identify potential overfitting, vulnerability to adversarial attacks, and neuron redundancy, which can inform decisions about neuron fusion for optimized performance.

## 3.3. Softermax function

When the bias term is disabled, the output layer scores are guaranteed to be non-negative, eliminating the necessity of applying the exponential function in the softmax operation. Based on this observation, we introduce Softermax, a simplified alternative to softmax.

The Softermax function provides a computationally efficient alternative to the standard softmax function, specifically tailored for scenarios where logits are non-negative. By replacing the exponentiation operation with a simple additive shift, Softermax reduces computational overhead while maintaining numerical stability and producing valid probability distributions.

Formally, for a vector of non-negative logits $\mathbf{x} = (x_1, x_2, \ldots, x_n)$ where $x_i \geq 0$, the Softermax function is defined as:

$$\text{Softermax}(x_i) = \frac{1 + x_i}{\sum_{j=1}^{n}(1 + x_j)}.$$

Here, $x_i$ represents the $i$-th logit. The additive shift $1 + x_i$ ensures numerical stability, particularly when logits are small or zero. The denominator normalizes the shifted logits, producing outputs that form a valid probability distribution, as they sum to 1.

Softermax is computationally efficient due to the elimination of exponentiation. Additionally, the function retains the relative scale of logits, offering an interpretable approximation of softmax behavior. To ensure well-defined outputs, inputs must satisfy $x_i \geq 0$, and logits are assumed to be non-negative real numbers ($\mathbb{R}^{+n}$).

This is particularly important for our application, as the defined product inherently produces only non-negative values.

## 3.4. 𝔼-Regularization

We propose a 𝔼-regularizer based on the 𝔼-product intra-similarity minimization used by Bouhsine et al for Anchor-Free Contrastive Learning (AFCL) [1]. This regularizer minimizes 𝔼-similarity score between weight vectors, encouraging intra-orthogonality.

## 3.5. 𝔼-ViT

Our 𝔼-ViT retains the original design but uses 𝔼-neurons instead, omitting any activation functions after the layer.

### 3.5.1 𝔼-MHA

Additionally, we replace the standard scaling factor $d_k$ with a learnable parameter that resembles the 𝔼-neuron scaling factor. And we use the defined Softermax function instead of the softmax.

$$\text{Attention}(Q, K, V) = \text{softermax}(\Theta \cdot QK^T)V \quad (15)$$

where $\Theta = \left(\frac{n}{\log(1+n)}\right)^\alpha$ and $\alpha$ is a learnable parameter. This adjustment enhances the flexibility of scaling within the attention mechanism.

### 3.5.2 Random Token Masking

Inspired by the Masked Autoencoder (MAE) approach [8], we apply random token masking between encoder blocks. Specifically, we randomly remove $p\%$ of input tokens and replace them with a single mask token. This strategy bolsters robustness and mitigates overfitting.

Let:
- $\mathbf{X} = [x_1, x_2, \ldots, x_n]$ be the input sequence,
- $p$ be the masking ratio, representing the probability of masking each token.

1. Binary Mask Generation:

For each token $x_i$, generate a binary mask $M_i$ such that:

$$M_i = \begin{cases} 1 & \text{with probability } p \\ 0 & \text{with probability } 1-p \end{cases}$$

2. Apply the Mask:

Define each masked token $x'_i$ as:

$$x'_i = M_i \cdot [\text{MASK}] + (1 - M_i) \cdot x_i$$

giving the masked sequence $\mathbf{X}' = [x'_1, x'_2, \ldots, x'_n]$, where $p \times n$ tokens are approximately replaced by [MASK].

## 4. Results

### 4.1. Experimental Setup

Our experimental framework compares the E-neuron and traditional neuron ensuring comparable parameter counts across all models using different image classification datasets: (1) E-neuron-based MLP architectures with Dropout; (2) a traditional MLP with ReLU activation, Dropout, Layer Normalization; (3) ViT [6] Architecture using E-neuron and traditional neuron.

The configuration for the ViT model [8, 28, 28] used in this study consists of a total of six layers. Each layer contains 128 hidden units. The model is designed with 512 parameters and utilizes two attention heads (**1.2m params**).

The results for the MLP and E-MLP (NMN) models uses the same patch-embedding scheme as the ViT, followed by global average pooling. We compare two models with the same number of neurons (**0.2m params**), but the E-neurons do not use any activation functions, while the traditional neurons have GeLU activation function [9]. For data augmentation we use Rotation, Flip Left-Right, Up-Down, and color jitter.

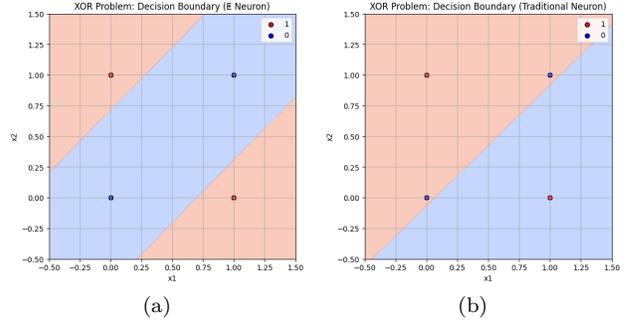

Figure 2. Decision boundaries of one trained (a) E-Neuron without scale (Code 8.4) vs (b) Traditional Neuron

To ensure the reproducibility of our results, all experiments are conducted on NVIDIA L4 GPUs using Jax/Flax [20] over GCP/Colab.

### 4.2. XOR Delima

One significant challenge in the field is the XOR problem, which typically requires nonlinear transformations or multiple layers in traditional neural networks to achieve accurate results.

As shown in Figure 2, a single E-neuron successfully resolves the XOR problem without the need for additional layers or non-homomorphic activation functions, demonstrating its capacity to handle nonlinear problems with minimal complexity.

### 4.3. Do you even MNIST bro?

In Figure 3, We manually placed 10 E-neurons (large circles) on the t-SNE plot of the MNIST dataset. We then used the maximum E-similarity score between the neurons and the t-SNE point of each image to predict the class. This approach achieved an accuracy of 73%.

### 4.4. Vision Classification Task

Table 1 presents the accuracy results across five benchmark image classification datasets, comparing architectures trained with traditional neurons and activation-free E-neurons in both MLP and Vision Transformer (ViT) configurations. Across all tasks, models incorporating E-neurons demonstrate improved or comparable accuracy over traditional neuron models trained for 200 epochs, underscoring the potential of E-neurons to deliver higher performance with fewer architectural complexities.

For CIFAR-10, the traditional ViT-t architecture achieves a test accuracy of 72.91%, while the E-

| | Traditional Neuron with GeLU Activation Function | | | | |
|---|---|---|---|---|---|
| Model | CIFAR10 [13] | CIFAR100 [13] | Caltech101 [15] | Oxford Flowers [23] | STL10 [3] |
| (in$^2$/# cls/steps) | $32^2$/10/390 | $32^2$/100 | $96^2$/102/23 | $224^2$/102/15 | $96^2$/10/78 |
| MLP | /4) ±25.31% | /4) 10.09% | /8) ±14.68% | /16) ±1.42% | /8) ±23.22 % |
| ViT-t | /4) ±72.91% | /4) ±36.93% | /8) ±30.21% | /16) ±31.22% | /8) ±49.73% |
| | Activation-Free E-Neuron | | | | |
| E-MLP | /4) ±47.36% | /4) ±25.14% | /8) ±24.15% | /16) ±20.20 % | /8) ±42.25 % |
| E-ViT-t | /4) ±74.22% | /4)±40.75% | /8) ±34.31% | /16) ±31.42% | /8)±51.95% |

Table 1. Comparison of test accuracy across multiple image classification datasets for architectures trained from scratch with traditional neurons and activation-free E-neurons. Results highlight the improved or comparable performance of E-neuron models, underscoring their effectiveness in both MLP and Vision Transformer (ViT) configurations

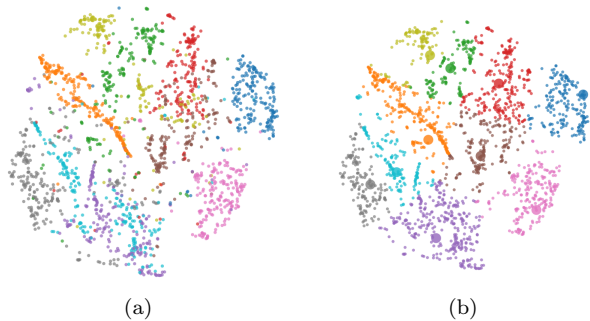

Figure 3. T-SNE plot of MNIST with (a) true labels and (b) predicted labels by manual fitting of 10 E-Neurons (large circles) without scale and 73% accuracy

ViT-t model outperforms it slightly at 74.22%, highlighting the effectiveness of E-neurons in enhancing generalization without additional activation functions. In CIFAR-100, E-neuron models similarly outperform their traditional counterparts, with the E-ViT-t model achieving 40.75% accuracy compared to 36.93% for the traditional ViT-t. These consistent improvements reflect the robustness of the E-neuron approach across datasets of varying complexity.

In MLP configurations, the advantage of E-neurons is further pronounced. The E-MLP model achieves 47.36% accuracy on CIFAR-10 and 25.14% on CIFAR-100, compared to the traditional MLP's 25.31% on CIFAR-10 and 10.09% on CIFAR-100. These results illustrate that E-neurons not only eliminate the need for explicit activation functions but also improve overall performance in simpler architectures, making them an efficient alternative for network design in resource-constrained settings.

On other datasets like Caltech101 and STL10, E-ViT-t models maintain their performance advantage, with respective accuracies of 34.31% and 51.95%, outperforming the traditional ViT-t's scores of 30.21% and 49.73%. The performance gains observed in E-neuron models across these datasets reinforce the generalizability and scalability of E-neuron-based architectures.

## 5. Discussion and Analysis of E-Neuron Performance

The results highlight the capability of E-neuron in achieving high accuracy without relying on traditional activation functions. The E-product introduces implicit non-linearity, capturing complex data patterns in a manner that preserves more information compared to standard activation-based approaches. This novel non-linear processing method represents a shift in how neural networks interpret data, reducing common information loss issues typically introduced by activation functions ( Table 1).

The E-neuron architecture enhances interpretability by embedding non-linearity directly within the E-product, thus preserving essential geometric relationships. This design enables a more intuitive understanding of neuron interactions and allows for straightforward visual representations of their behavior.

Additionally, E-neurons circumvent several stability challenges seen with traditional activation functions. Standard non-linearities can lead to gradient saturation and "dead" neurons [25], with saturating functions often resulting in vanishing gradients, while non-saturating ones can cause exploding gradients or neuron death, requiring added techniques like batch normalization and gradient clipping. The E-product avoids these issues by maintaining non-saturating internal non-linearity, offering stable training dynamics, and reducing the dependency on additional regularization techniques.

### 5.1. Artificial Neurons that Matter

The E-product draws intriguing analogies with physical laws, such as the inverse-square law, hinting at a potential new paradigm for understanding neural networks. Here, neurons are not merely linear separators but op-erate as geometric constructs within a non-linear manifold. This interpretation reimagines neural network layers as inherently physical, aligning closer to natural principles. Such a view opens pathways to network architectures that are both adaptable and more naturally aligned with fundamental scientific laws. Furthermore, including a learnable scale parameter mitigates the risk of weight explosion. Without this parameter, weights grow excessively, leading to instability.

### 5.2. Effect of 𝔼-Regularization on Output Layer Representation

Our examination of well-trained model weight matrices shows a tendency toward orthogonality among neurons, which supports improved generalization and robustness. Non-orthogonality or linear dependencies in the weight matrix often indicate suboptimal training. To promote orthogonality, The 𝔼-regularizer encourages orthogonality among weight vectors, preventing neuron representations from collapsing into similar configurations. This effect is especially beneficial for distinguishing similar classes, such as "cats" versus "dogs" in datasets like CIFAR-10. Figure 4 illustrates how the 𝔼-regularizer prevents neuron collapse in the output layer, as shown through NMS plots, providing distinct and separated representations for each class compared to training without regularization which leads to the neuron misclassification of the classes dog vs cats as the neuron weights of those two classes are highly similar.

## 6. Limitations and Future Directions

While the 𝔼-neuron architecture introduces several advantages, certain limitations may affect its scalability and compatibility across different applications.

The 𝔼-product does not satisfy standard associative and distributive properties, fundamental to many matrix operations in deep learning. This limitation could restrict the 𝔼-neuron's adaptability, especially in code optimizations and architectures that depend on these algebraic properties for performance and flexibility.

As for the computational efficiency, traditional neurons compute a dot product followed by a ReLU activation, requiring about $2d + 1$ FLOPs per neuron (where $d$ is the input dimension). In contrast, 𝔼-neurons use the 𝔼-product, which includes a squared Euclidean distance and magnitude normalization, totaling approximately $5d - 1$ FLOPs per neuron.

To analyze the computational overhead, we calculate the FLOP ratio between 𝔼-neurons and traditional neurons:

$$\text{Efficiency Ratio} = \frac{5d - 1}{2d + 1} \approx 2.5$$

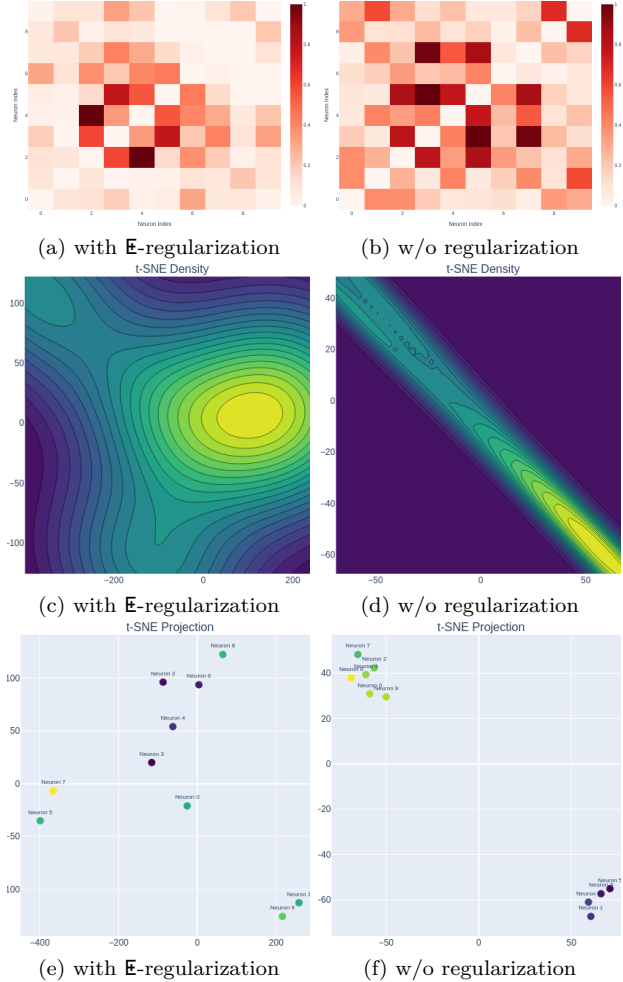

Figure 4. 𝔼-reguliarization prevent the neuron collapse between Dog (idx 5) and Cat (idx 3) Neurons in the output layer as shown by the 𝔼-similarity matrix (a,b) and NMS Plots (c,d,e,f)

This suggests that 𝔼-neurons require approximately 2.5 times the FLOPs of conventional neurons. However, the 𝔼-neuron's design eliminates the need for separate activation functions, potentially streamlining the overall network structure and reducing the memory load in deeper architectures.

The absence of activation functions between layers in the 𝔼-neuron architectures results in more information-preserving representations, yet this can lead to a higher risk of overfitting.

Drawing an analogy to Hebbian learning, which states "neurons that fire together, wire together," [18] we observe that overfitting in neurons can manifest as neuron collapse. This collapse occurs when multiple neurons converge to encode identical information, occupying the same region in the neural space. Figure 5 illustrates this phenomenon using Neural-Matter State (NMS) plots revealing a collapsed distribution indicating overfitting in the output layer.

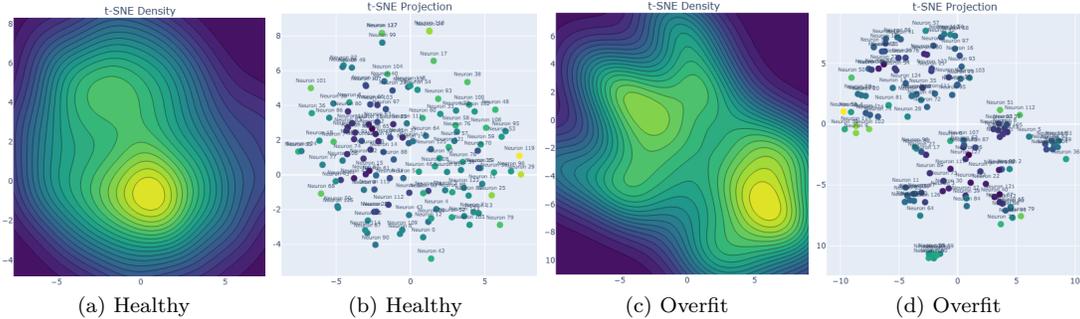

(a) Healthy     (b) Healthy     (c) Overfit     (d) Overfit

Figure 5. Analysis of the neurons in the third layer of an MLP head trained on CIFAR-10, using a Neural-Matter State (NMS) plot, reveals: (a, b) a well-fitting, healthy distribution of neuron specialization and (c, d) overfitting appears as a collapsed distribution.

Future research should concentrate on comprehending this new paradigm and developing architectures specifically optimized for this new neuron.

## 7. Related Works

### 7.1. Inverse-Square Laws

The inverse-square law describes a principle where a specified physical quantity or intensity is inversely proportional to the square of the distance from the source. This relationship is foundational in various fields, particularly in physics, and underpins several fundamental laws [12]. Newton's Law of Universal Gravitation is one of the earliest applications of the inverse-square law. It states that the gravitational force between two masses is inversely proportional to the square of the distance between their centers [22]. Similarly, Coulomb's Law describes the electrostatic force between two charged particles, with a force that diminishes with the square of the distance [4]. Gauss's Law, though formulated differently, implies an inverse-square relationship for electric and gravitational flux through a closed surface, connecting field flux to the source charge or mass within that surface [7].

### 7.2. Multi Layer Perceptron Architectures

Recent research in computer vision explores Multilayer Perceptrons (MLPs) and Vision Transformers [6] as alternatives to conventional convolutional models, focusing on simplicity and computational efficiency. Models like MLP-Mixer [30] and gMLP [17] have demonstrated competitive results on benchmarks like ImageNet by using operations such as matrix multiplication and spatial gating, effectively capturing spatial information without self-attention.

## 8. Conclusion

Perhaps artificial intelligence's greatest limitation has been our stubborn fixation on the human brain as the pinnacle of intelligence. The universe itself, governed by elegant and powerful laws, demonstrates intelligence far beyond human cognition. These fundamental laws - which shape galaxies and guide quantum particles - represent a deeper form of intelligence that we have largely ignored in our pursuit of AI.

In this paper, we broke free from biological constraints by drawing direct inspiration from inverse-square law [12], Coulomb's law [4], Newton's first law of motion [22], Gauss's Law [7], and Bouhsine's Products [1]. By redefining neural fundamentals through the lens of physics and abstract topology, we demonstrated that a single activation-free neuron could solve the XOR problem - a task that traditionally required multiple neurons with complex activation functions. Our extensive experiments show that networks built with the Ǝ-neurons consistently outperform traditional architectures without using any non-linear activation functions, suggesting that we have only scratched the surface of what's possible when we look beyond biological metaphors.

### License

The source code, algorithms, and all contributions presented in this work are licensed under the GNU Affero General Public License (AGPL) v3.0. This license ensures that any use, modification, or distribution of the code and any adaptations or applications of the underlying models and methods must be made publicly available under the same license. This applies whether the work is used for personal, academic, or commercial purposes, including services provided over a network.

## Acknowledgment

The Google Developer Expert program and Google AI/ML Developer Programs team supported this work by providing Google Cloud Credit. I want to extend my gratitude to the staff at High Grounds Coffee Roasters for their excellent coffee and peaceful atmosphere. I would also like to thank Dr. Andrew Ng for creating the Deep Learning course that introduced me to this field, without his efforts to democratize access to knowledge, this work would not have been possible. Additionally, I want to express my appreciation to all the communities I have been part of, especially MLNomads, Google Developers, and MLCollective communities.

# Appendix

## 8.1. Geometric Topology

In geometric topology, various types of spaces are used to study notions of distance and convergence. Each type of space has a specific set of properties defined by a function known as a metric or a generalization thereof. We describe below metric spaces, semi-metric spaces, and pseudo-metric spaces, emphasizing the differences in their definitions.

### 8.1.1 Metric Space

A metric space is a set $X$ equipped with a distance function $d : X \times X \to \mathbb{R}$, called a metric, which satisfies the following properties for all $x, y, z \in X$:

1. Non-negativity: $d(x, y) \geq 0$.
2. Identity of indiscernibles: $d(x, y) = 0$ if and only if $x = y$.
3. Symmetry: $d(x, y) = d(y, x)$.
4. Triangle inequality: $d(x, z) \leq d(x, y) + d(y, z)$.

These properties ensure that a metric space has a well-defined notion of distance between any pair of points in $X$, which is fundamental to many topological and analytical concepts.

### 8.1.2 Semi-Metric Space

A semi-metric space generalizes a metric space by relaxing the triangle inequality requirement. A semi-metric space is defined as a set $X$ with a distance function $d : X \times X \to \mathbb{R}$ that satisfies:

1. Non-negativity: $d(x, y) \geq 0$.
2. Identity of indiscernibles: $d(x, y) = 0$ if and only if $x = y$.
3. Symmetry: $d(x, y) = d(y, x)$.

In this case, $d$ provides a notion of distance that is symmetric and non-negative, though it does not necessarily satisfy the triangle inequality.

### 8.1.3 Pseudo-Metric Space

A pseudo-metric space is another generalization of a metric space, where the identity of indiscernibles requirement is omitted. Thus, a pseudo-metric space is a set $X$ with a distance function $d : X \times X \to \mathbb{R}$ that satisfies:

1. Non-negativity: $d(x, y) \geq 0$.
2. Symmetry: $d(x, y) = d(y, x)$.
3. Triangle inequality: $d(x, z) \leq d(x, y) + d(y, z)$.

In a pseudo-metric space, $d(x, y) = 0$ does not necessarily imply that $x = y$; points can have zero distance between them without being identical, which makes pseudo-metric spaces useful in contexts where such indistinguishability is needed.

### 8.1.4 $\mathbb{E}$ Propreties

The $\mathbb{E}$-product is pseudo-metric and $\overline{\mathbb{E}}$-product is semi-metric (Theorem 8.1), satisfying closure and commutativity, though lacking other conventional algebraic properties such as associativity, distributivity, and the existence of an identity element. This unique structure enables the proposed products to capture aspects of linearly projected vector (Unbounded) relationships that standard similarity metrics like the dot product, Euclidean distance, and Cosine Similarity may overlook (Table 2) especially in the created embedding space, where ntions like similarity, distance, parallalism, orthogonality lost meaning because of the use of non-linear functions, and the dot product that doesn't respect the rules of metric space.

Traditional similarity measures, such as the dot product, often fall short in capturing comprehensive vector relationships, especially when vectors share the same direction but vary in magnitude.

Consider a set of vectors $(1, 1)$, $(2, 2)$, $(3, 3)$, $(4, 4)$, $(5, 5)$, $(8, 8)$, and $(9, 9)$, which are all parallel and thus point in the same direction. For a new point $(6, 6)$, using cosine similarity (or dot product) would yield identical results for all these vectors since cosine similarity emphasizes direction while ignoring magnitude. This outcome is misleading, as intuitively, the point $(6, 6)$ is closest to $(5, 5)$ in both direction and magnitude. By disregarding magnitude, the dot product and cosine similarity fail to differentiate the proximity between $(6, 6)$ and the individual vectors.

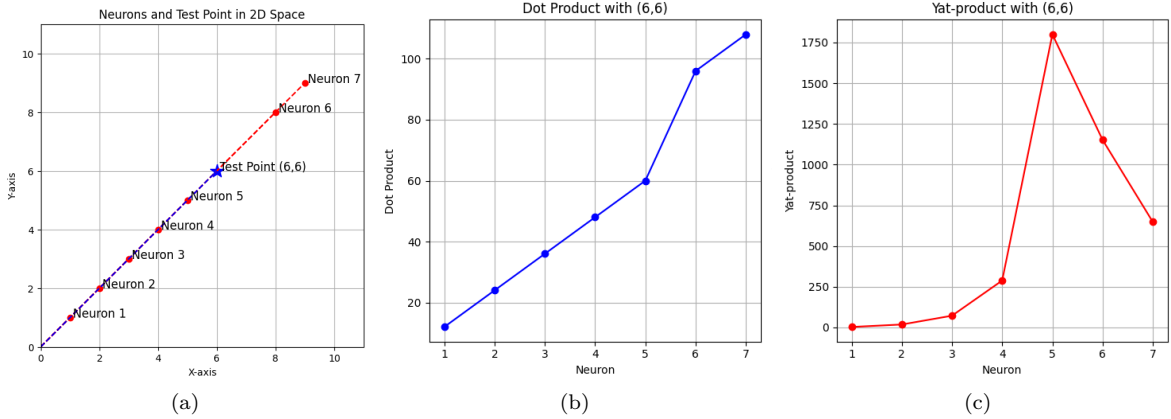

(a)            (b)            (c)

Figure 6. Comparison of similarity measurements for the test point $(6, 6)$ with vectors using (a) the dot product and (b) the ᴇ-product (Code 8.3). In (b), the dot product scales with vector magnitude, often exaggerating similarity based on size alone. In (c), the ᴇ-product more accurately reflects the relative magnitude and distance, correctly identifying $(5, 5)$ as the closest match to $(6, 6)$.

The ᴇ-product addresses this limitation by incorporating both magnitude and distance information. Unlike the dot product, which is primarily based on angular similarity, the ᴇ-product is designed to account for both directional alignment and the relative distances between vectors. When applying the ᴇ-product to compare $(6, 6)$ with each neuron vector, it correctly identifies $(5, 5)$ as the closest match, offering a nuanced understanding that aligns with the intuitive notion of similarity.

Figure 6 visually demonstrates the differences between the dot product and the ᴇ-product. Plot (b) shows the dot product's tendency to favor larger vector magnitudes. In contrast, the ᴇ-product plot (c) effectively differentiates between vectors based on both magnitude and spatial proximity, underscoring its advantage in applications that require a holistic similarity measure.

As for the computational efficiency, a dot product requires about $2d$ FLOPs per operation (where $d$ is the input dimension). In contrast, ᴇ-neurons use the ᴇ-product, which includes a squared Euclidean distance and magnitude normalization, totaling approximately $5d - 1$ FLOPs per operation.

To analyze the computational overhead, we calculate the FLOP ratio between ᴇ-neurons and traditional neurons:

$$\text{Efficiency Ratio} = \frac{5d - 1}{2d}$$

This suggests that ᴇ-product requires approximately 2.5 times the FLOPs of dot product, similarly for $\overline{\text{ᴇ}}$.

## 8.2. ᴇ is pseudo-metric space

**Theorem 8.1** ($\overline{\text{ᴇ}}$ is semi-metric and ᴇ is pseudo-metric)**.** *Let* $(\mathbb{R}^n, \overline{\text{ᴇ}})$ *and* $(\mathbb{R}^n, \text{ᴇ})$ *be two spaces where:*

$$\text{ᴇ}(e_i, e_j) = \frac{o_{ij}}{d_{ij}^2} = \frac{(e_i \cdot e_j)^2}{||e_i - e_j||^2}$$

$$\overline{\text{ᴇ}}(e_i, e_j) = \frac{d_{ij}^2}{o_{ij}} = \frac{||e_i - e_j||^2}{(e_i \cdot e_j)^2}$$

*for* $e_i, e_j \in \mathbb{R}^n \setminus \{0\}$ *where* $e_i \neq e_j$, *with* $d_{ij}^2 = |e_i - e_j|^2$ *being the squared Euclidean distance and* $o_{ij} = (e_i \cdot e_j)^2$ *being the squared dot product.*

*Then* $(\mathbb{R}^n, \overline{\text{ᴇ}})$ *is semi-metric, while* $(\mathbb{R}^n, \text{ᴇ})$ *is pseudo-metric.*

*Proof.* We structure this proof into four parts:

| Property | · | ‖.‖ | Cosine | $\mathbb{E}$ | $\bar{\mathbb{E}}$ |
|---|---|---|---|---|---|
| Non-negativity | x | ✓ | x | ✓ | ✓ |
| Identity of Indiscernibles | x | ✓ | x | x | ✓ |
| Symmetry | ✓ | ✓ | ✓ | ✓ | ✓ |
| Triangle Inequality | x | ✓ | x | ✓ | x |
| Space Type |  | Metric |  | pseudo-metric | pseudo-metric |
| Orthogonality | ✓ | No | ✓ | ✓ | ✓ |
| Distance | X | ✓ | X | ✓ | ✓ |
| Probability | No | No | No | No | No |
| FLOPS | $d$ | $3d$ | $4d+1$ | $5d-1$ | $5d-1$ |
| Bounds | $]-\infty, \infty[$ | $(0,\infty[$ | $[-1, 1]$ | $(0, \infty[$ | $(0, \infty[$ |

Table 2. Comparison of Different Metrics

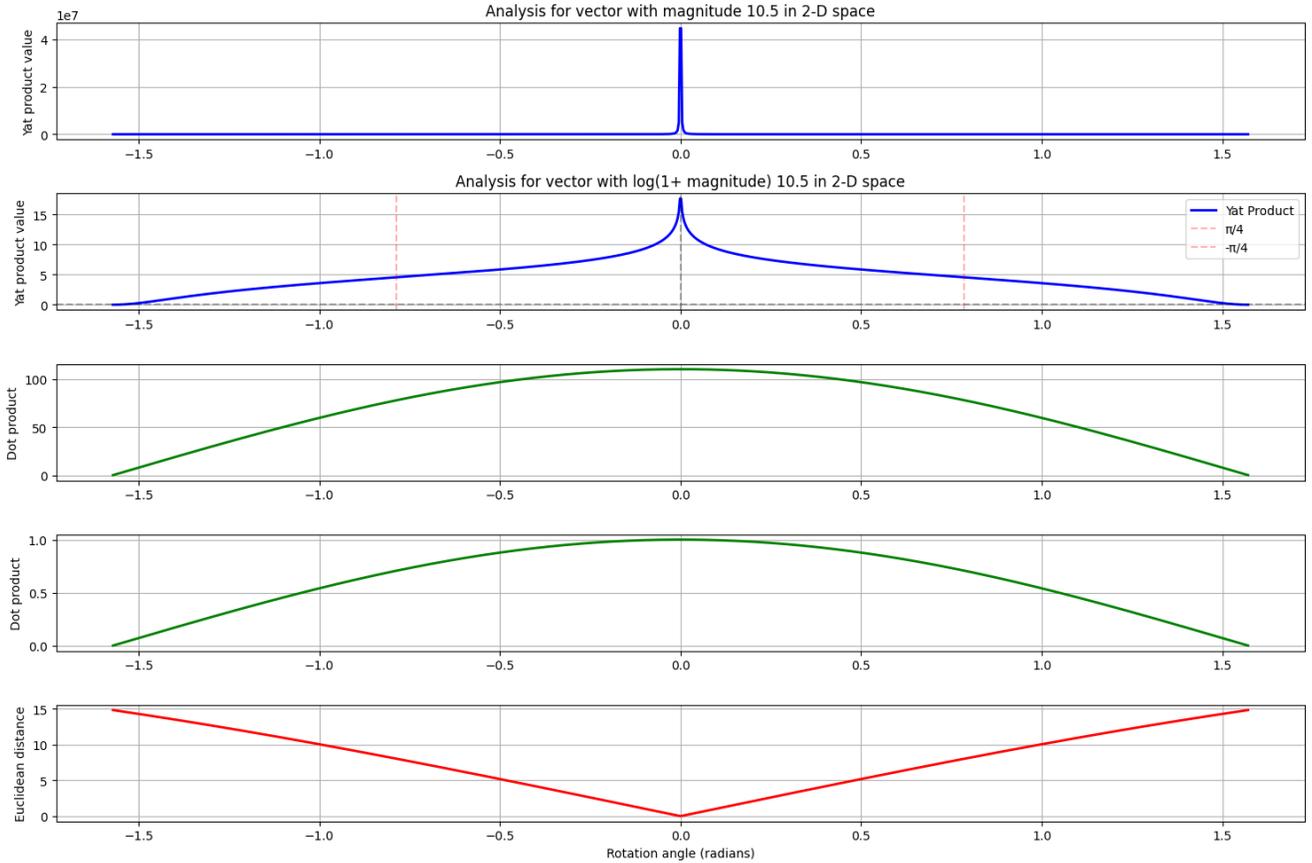

Figure 7. A comparison of $\mathbb{E}$, ·, and ‖ . ‖ on vectors with a magnitude of 10, alongside other vectors with the same magnitude that span from $-\pi/2$ to $\pi/2$ in relation to the main vector.

1. Preliminary observations and domain analysis
2. Proof of common properties for both measures
3. Proof that $\mathbb{E}$ is a pseudo-metric
4. Proof that $\bar{\mathbb{E}}$ is a semi-metric

**Part I: Preliminary Observations**

Before proving the metric properties, we must establish the domain where these measures are well-defined:

1. For non-zero vectors $e_i, e_j$:

- $d_{ij}^2 = 0 \iff e_i = e_j$
- $o_{ij} = 0 \iff e_i \perp e_j$ (vectors are orthogonal)

2. Domain restrictions:
- $\mathbb{E}$ is defined when $d_{ij}^2 \neq 0$ (distinct vectors)
- $\overline{\mathbb{E}}$ is defined when $o_{ij} \neq 0$ (non-orthogonal vectors)

**Part II: Common Properties**

Both measures satisfy the following properties:

**1. Non-negativity:** Since both $d_{ij}^2$ and $o_{ij}$ are squared quantities:

$$d_{ij}^2 = ||e_i - e_j||^2 \geq 0 \quad \text{and} \quad o_{ij} = (e_i \cdot e_j)^2 \geq 0$$

Therefore:

$$\mathbb{E}(e_i, e_j) \geq 0 \quad \text{and} \quad \overline{\mathbb{E}}(e_i, e_j) \geq 0$$

**2. Identity of Indiscernibles:** For both measures, we prove this bidirectionally:

($\Rightarrow$) If $e_i = e_j$:
- $d_{ij}^2 = 0$
- $o_{ij} = ||e_i||^4 > 0$ (for non-zero vectors)

Therefore, $\mathbb{E}(e_i, e_j) = 0$ and $\overline{\mathbb{E}}(e_i, e_j) = 0$

($\Leftarrow$) If $\mathbb{E}(e_i, e_j) = 0$ or $\overline{\mathbb{E}}(e_i, e_j) = 0$:
- For $\mathbb{E}$: $\frac{o_{ij}}{d_{ij}^2} = 0 \implies o_{ij} = 0$ (since $d_{ij}^2 \neq 0$ for distinct vectors)
- For $\overline{\mathbb{E}}$: $\frac{d_{ij}^2}{o_{ij}} = 0 \implies d_{ij}^2 = 0$ (since $o_{ij} \neq 0$ in domain)

In both cases, this implies that this rule stand for $\overline{\mathbb{E}}$, but not for $\mathbb{E}$.

**3. Symmetry:** Symmetry follows from the symmetry of dot product and Euclidean distance:

$$\mathbb{E}(e_i, e_j) = \frac{(e_i \cdot e_j)^2}{||e_i - e_j||^2} = \frac{(e_j \cdot e_i)^2}{||e_j - e_i||^2} = \mathbb{E}(e_j, e_i)$$

And similarly for $\overline{\mathbb{E}}$.

**Part III: Proof the triangle inequality for $\mathbb{E}$**

To prove $\mathbb{E}$ satisfies the triangle inequality, we proceed in steps:

Given:
$$\mathbf{e}_1 \mathbb{E} \mathbf{e}_2 = \frac{(\mathbf{e}_1 \cdot \mathbf{e}_2)^2}{||\mathbf{e}_2 - \mathbf{e}_1||^2}.$$

Let:
- $\mathbf{e}_1$ and $\mathbf{e}_2$ be vectors in $\mathbb{R}^n$,
- $\theta$ be the angle between $\mathbf{e}_1$ and $\mathbf{e}_2$.

The dot product between $\mathbf{e}_1$ and $\mathbf{e}_2$ can be written as:

$$\mathbf{e}_1 \cdot \mathbf{e}_2 = ||\mathbf{e}_1|| \, ||\mathbf{e}_2|| \cos \theta.$$

Thus, $(\mathbf{e}_1 \cdot \mathbf{e}_2)^2$ becomes:

$$(\mathbf{e}_1 \cdot \mathbf{e}_2)^2 = (||\mathbf{e}_1|| \, ||\mathbf{e}_2|| \cos \theta)^2 = ||\mathbf{e}_1||^2 \, ||\mathbf{e}_2||^2 \cos^2 \theta.$$

The Euclidean distance between $\mathbf{e}_1$ and $\mathbf{e}_2$ is:

$$||\mathbf{e}_2 - \mathbf{e}_1||^2 = ||\mathbf{e}_1||^2 + ||\mathbf{e}_2||^2 - 2 \, ||\mathbf{e}_1|| \, ||\mathbf{e}_2|| \cos \theta.$$

Now we substitute these expressions into the formula for $\mathbf{e}_1 \mathbb{E} \mathbf{e}_2$:

$$\mathbf{e}_1 \mathbb{E} \mathbf{e}_2 = \frac{||\mathbf{e}_1||^2 \, ||\mathbf{e}_2||^2 \cos^2 \theta}{||\mathbf{e}_1||^2 + ||\mathbf{e}_2||^2 - 2 \, ||\mathbf{e}_1|| \, ||\mathbf{e}_2|| \cos \theta}.$$

Let's simplify by defining:

- $A = ||\mathbf{e}_1||$,
- $B = ||\mathbf{e}_2||$.

Thus, the expression becomes:

$$f(\theta) = \mathbf{e}_1 \mathbb{E} \mathbf{e}_2 = \frac{A^2 B^2 \cos^2 \theta}{A^2 + B^2 - 2AB \cos \theta}.$$

Let's factor out common terms in the numerator. Notice that each term in the numerator has a factor of $A^2 B^2 \sin \theta$, so we can factor that out:

$$f'(\theta) = \frac{A^2 B^2 \sin \theta \left[ -2 \cos \theta (A^2 + B^2 - 2AB \cos \theta) - 2AB \cos^2 \theta \right]}{(A^2 + B^2 - 2AB \cos \theta)^2}.$$

Now, distribute $-2 \cos \theta$ in the first term inside the brackets:

$$= \frac{A^2 B^2 \sin \theta \left[ -2A^2 \cos \theta - 2B^2 \cos \theta + 4AB \cos^2 \theta - 2AB \cos^2 \theta \right]}{(A^2 + B^2 - 2AB \cos \theta)^2}.$$

Combine the $\cos^2 \theta$ terms:

$$= \frac{A^2 B^2 \sin \theta \left[ -2A^2 \cos \theta - 2B^2 \cos \theta + 2AB \cos^2 \theta \right]}{(A^2 + B^2 - 2AB \cos \theta)^2}.$$

Thus, the simplified form of $f'(\theta)$ is:

$$f'(\theta) = \frac{-2A^2 B^2 \sin \theta \left( A^2 \cos \theta + B^2 \cos \theta - AB \cos^2 \theta \right)}{(A^2 + B^2 - 2AB \cos \theta)^2}.$$

This form is simpler and allows us to see that the sign of $f'(\theta)$ depends on the sign of $-\sin \theta$, which is non-positive on the interval $[0, \pi]$. Therefore, $f'(\theta) \leq 0$ on this interval, confirming that $f(\theta)$ is monotonically decreasing.

Since $f(\theta)$ is monotonically decreasing, it follows that $\mathbb{E}(\mathbf{e}_1, \mathbf{e}_2) = f(\theta)$ decreases as $\theta$ increases.

Applying the Angular Triangle Inequality Angles in Euclidean space satisfy the triangle inequality (Cauchy–Schwarz inequality):

$$\theta_{ik} \leq \theta_{ij} + \theta_{jk}.$$

Since $\mathbb{E}(\mathbf{e}_i, \mathbf{e}_j)$ is a decreasing function of $\theta$, we conclude:

$$\mathbb{E}(\mathbf{e}_i, \mathbf{e}_k) \leq \mathbb{E}(\mathbf{e}_i, \mathbf{e}_j) + \mathbb{E}(\mathbf{e}_j, \mathbf{e}_k).$$

Since $\mathbb{E}(\mathbf{e}_i, \mathbf{e}_j) = \frac{1}{\overline{\mathbb{E}}(\mathbf{e}_i, \mathbf{e}_j)}$, we deduce that the $\overline{\mathbb{E}}$ is a increasing function of $\theta$, we conclude:

$$\overline{\mathbb{E}}(\mathbf{e}_i, \mathbf{e}_k) \leq \overline{\mathbb{E}}(\mathbf{e}_i, \mathbf{e}_j) + \overline{\mathbb{E}}(\mathbf{e}_j, \mathbf{e}_k).$$

, the only problem that is preventing the $\overline{\mathbb{E}}$ from being a fully metric space is not being defined when $e_i \perp e_j$, but we can remedy this with an $\epsilon$ so now it becomes $\overline{\mathbb{E}} = \frac{||\mathbf{e}_2 - \mathbf{e}_1||^2}{(\mathbf{e}_1 \cdot \mathbf{e}_2)^2 + \epsilon}$, we do the same thing to the $\mathbb{E}$ to define it when $e_i = e_j$, just unlike the $\overline{\mathbb{E}}$, this operation doesn't change the fact that $\mathbb{E}$ remains pseudo-metric. $\square$

## 8.3. $\mathbb{E}$-product vs Dot Product

```
# Plot the neurons and the test point in the 2D space

plt.figure(figsize=(6, 6))

# Plot neurons as red dots
for i, neuron in enumerate(neurons):
    plt.plot(neuron[0], neuron[1], 'ro')
    plt.text(neuron[0] + 0.1, neuron[1], f'Neuron {i+1}', fontsize=12)
```

```python
# Plot the test point (6, 6) as a blue star
plt.plot(test_point[0], test_point[1], 'b*', markersize=15, label='Test Point (6,6)')
plt.text(test_point[0] + 0.1, test_point[1], 'Test Point (6,6)', fontsize=12)

# Set up the plot limits and labels
plt.xlim(0, 7)
plt.ylim(0, 7)
plt.xlabel('X-axis')
plt.ylabel('Y-axis')
plt.title('Neurons and Test Point in 2D Space')

# Draw lines connecting the origin to the neurons and test point
for neuron in neurons:
    plt.plot([0, neuron[0]], [0, neuron[1]], 'r--')

plt.plot([0, test_point[0]], [0, test_point[1]], 'b--')

plt.grid(True)
plt.gca().set_aspect('equal', adjustable='box')
plt.show()

%%%%%% TODO change this code in the paper to the new one
import numpy as np
import matplotlib.pyplot as plt

# Define the neurons as vectors
neurons = np.array([[1, 1], [2, 2], [3, 3], [4, 4], [5, 5], [7, 7], [8, 8]])
test_point = np.array([6, 6])

# dot product function
def cosine_similarity(v1, v2):
    dot_product = np.dot(v1, v2)
    return dot_product

# Yat-product function
def yat_product(v1, v2, epsilon=1e-6):
    dot_product_squared = np.dot(v1, v2) ** 2
    distance_squared = np.linalg.norm(v2 - v1) ** 2
    return dot_product_squared / (distance_squared + epsilon)
# Calculate cosine similarities and yat-products
cosine_similarities = [cosine_similarity(test_point, neuron) for neuron in neurons]
yat_products = [yat_product(test_point, neuron) for neuron in neurons]

# Plot the cosine similarities
plt.figure(figsize=(10, 5))

plt.subplot(1, 2, 1)
plt.plot(range(1, 6), cosine_similarities, marker='o', color='blue', label='Cosine Similarity
plt.title("Cosine Similarity with (6,6)")
plt.xlabel("Neuron")
plt.ylabel("Cosine Similarity")
plt.ylim(0, 1.1)
```

```python
plt.grid(True)
plt.xticks([1, 2, 3, 4, 5])

# Plot the Yat-products
plt.subplot(1, 2, 2)
plt.plot(range(1, 6), yat_products, marker='o', color='red', label='Yat-product')
plt.title("Yat-product with (6,6)")
plt.xlabel("Neuron")
plt.ylabel("Yat-product")
plt.grid(True)
plt.xticks([1, 2, 3, 4, 5])

plt.tight_layout()
plt.show()
```

### E-Neuron with numpy

```python
import numpy as np
def yat_neuron(X, w, b):
    # Squared dot product
    dot_squared = np.dot(X, w) ** 2
    # Squared Euclidean distance
    distance_squared = np.sum((w - X) ** 2, axis=1)
    # Avoid division by zero by adding a small epsilon
    epsilon = 1e-6
    return dot_squared / (distance_squared + epsilon) + b
```

### 8.4. XOR Code

```python
import numpy as np
np.random.seed(42)  # For reproducibility
w = np.random.randn(2)  # Random weights for a 2D input
b = np.random.randn()   # Random bias
# XOR dataset: inputs and corresponding outputs
X = np.array([[0, 0], [0, 1], [1, 0], [1, 1]])  # Inputs
y = np.array([0, 1, 1, 0])  # Expected outputs for XOR

# Apply custom neuron to each input in XOR dataset
outputs = yat_neuron(X, w, b)

# Print results
print(w)
print(b)
print(outputs)
w, b, outputs

from scipy.optimize import minimize
# Loss function: Mean Squared Error between the neuron output and the target XOR output
def loss_function(params):
    w = params[:2]  # First two values are weights
    b = params[2]   # Last value is bias
    outputs = yat_neuron(X, w, b)
    return np.mean((outputs - y) ** 2)  # Mean Squared Error (MSE)
```

```python
# Initial parameters: [w1, w2, b]
initial_params = np.append(w, b)

# Optimize the weights and bias using 'minimize' from SciPy
result = minimize(loss_function, initial_params, method='BFGS')

# Extract optimized weights and bias
optimized_params = result.x
optimized_weights = optimized_params[:2]
optimized_bias = optimized_params[2]

# Apply optimized weights and bias to XOR dataset
optimized_outputs = yat_neuron(X, optimized_weights, optimized_bias)

print('################')
print(optimized_weights)
print(optimized_bias)
print(optimized_outputs)

import matplotlib.pyplot as plt

# Function to plot XOR data and decision boundary
def plot_xor_decision_boundary(weights, bias):
    # XOR input data points
    X_pos = X[y == 1]  # Points where y == 1
    X_neg = X[y == 0]  # Points where y == 0

    # Create a meshgrid for the plot
    xx, yy = np.meshgrid(np.linspace(-0.5, 1.5, 400), np.linspace(-0.5, 1.5, 400))
    grid_points = np.c_[xx.ravel(), yy.ravel()]

    # Compute neuron output for each point in the grid
    Z = yat_neuron(grid_points, weights, bias)
    Z = Z.reshape(xx.shape)
    print(Z)
    print(xx)
    # Plot the decision boundary and XOR points
    plt.figure(figsize=(6, 6))
    plt.contourf(xx, yy, Z > 0.5, alpha=0.5, cmap='coolwarm')  # Decision boundary
    plt.scatter(X_pos[:, 0], X_pos[:, 1], color='red', label='1', edgecolors='k')
    plt.scatter(X_neg[:, 0], X_neg[:, 1], color='blue', label='0', edgecolors='k')
    plt.title("XOR Problem: Decision Boundary ( Neuron)")
    plt.xlabel('x1')
    plt.ylabel('x2')
    plt.legend()
    plt.grid(True)
    plt.show()

# Plot the XOR decision boundary with optimized weights and bias
plot_xor_decision_boundary(optimized_weights, optimized_bias)
```

### 8.5. Ɛ-Layer (NML) with Flax (github/mlnomadpy/nmn)

```python
import jax.numpy as jnp
from flax.linen.dtypes import promote_dtype
from flax.linen.module import Module, compact
from flax.typing import (
  PRNGKey as PRNGKey,
  Shape as Shape,
  DotGeneralT,
)

from typing import (
  Any,
)
import jax.numpy as jnp
import jax.lax as lax
from flax.linen import Module, compact
from flax import linen as nn
from flax.linen.initializers import zeros_init, lecun_normal
from typing import Any, Optional

class YatDense(Module):
    """
    Attributes:
      features: the number of output features.
      use_bias: whether to add a bias to the output (default: True).
      dtype: the dtype of the computation.
      param_dtype: the dtype passed to parameter initializers (default: float32).
      precision: numerical precision of the computation.
      kernel_init: initializer function for the weight matrix.
      bias_init: initializer function for the bias.
      epsilon: small constant.
    """
    features: int
    use_bias: bool = True
    dtype: Optional[Any] = None
    param_dtype: Any = jnp.float32
    precision: Any = None
    kernel_init: Any = nn.initializers.orthogonal()
    bias_init: Any = zeros_init()

    # Initialize alpha to 1.0
    alpha_init: Any = lambda key, shape, dtype: jnp.ones(shape, type)
    epsilon: float = 1e-6
    dot_general: DotGeneralT | None = None
    dot_general_cls: Any = None
    return_weights: bool = False

    @compact
    def __call__(self, inputs: Any) -> Any:
        """
        Args:
          inputs: The nd-array to be transformed.

        Returns:
```

```python
        The transformed input.
    """
    kernel = self.param(
        'kernel',
        self.kernel_init,
        (self.features, jnp.shape(inputs)[-1]),
        self.param_dtype,
    )
    alpha = self.param(
        'alpha',
        self.alpha_init,
        (1,),  # Single scalar parameter
        self.param_dtype,
    )
    if self.use_bias:
        bias = self.param(
            'bias', self.bias_init, (self.features,), self.param_dtype
        )
    else:
        bias = None

    inputs, kernel, bias = promote_dtype(inputs, kernel, bias, dtype=self.dtype)
    # Compute dot product between input and kernel
    if self.dot_general_cls is not None:
      dot_general = self.dot_general_cls()
    elif self.dot_general is not None:
      dot_general = self.dot_general
    else:
      dot_general = lax.dot_general
    y = dot_general(
      inputs,
      jnp.transpose(kernel),
      (((inputs.ndim - 1,), (0,)), ((), ())),
      precision=self.precision,
    )
    inputs_squared_sum = jnp.sum(inputs**2, axis=-1, keepdims=True)
    kernel_squared_sum = jnp.sum(kernel**2, axis=-1)
    distances = inputs_squared_sum + kernel_squared_sum - 2 * y

    # Element-wise operation
    y = y ** 2 / (distances + self.epsilon)
    scale = (jnp.sqrt(self.features) / jnp.log(1 + self.features)) ** alpha
    y = y * scale
    # Normalize y
    if bias is not None:
        y += jnp.reshape(bias, (1,) * (y.ndim - 1) + (-1,))
    if self.return_weights:
        return y, kernel
    return y
```

### 8.6. E-Layer (NML) with pytorch (github/mlnomadpy/nmn)

```python
import torch
```

```python
import torch.nn as nn
import torch.nn.functional as F
import math

class YatDense(nn.Module):
    """
    A PyTorch implementation of the Yat neuron with squared Euclidean distance transformation.

    Attributes:
        in_features (int): Size of each input sample
        out_features (int): Size of each output sample
        use_bias (bool): Whether to add a bias to the output
        dtype (torch.dtype): Data type for computation
        epsilon (float): Small constant to avoid division by zero
        kernel_init (callable): Initializer for the weight matrix
        bias_init (callable): Initializer for the bias
        alpha_init (callable): Initializer for the scaling parameter
    """
    def __init__(
        self,
        in_features: int,
        out_features: int,
        use_bias: bool = False,
        dtype: torch.dtype = torch.float32,
        epsilon: float = 1e-6,
        kernel_init: callable = None,
        bias_init: callable = None,
        alpha_init: callable = None
    ):
        super().__init__()

        # Store attributes
        self.in_features = in_features
        self.out_features = out_features
        self.use_bias = use_bias
        self.dtype = dtype
        self.epsilon = epsilon

        # Weight initialization
        if kernel_init is None:
            kernel_init = nn.init.xavier_normal_

        # Create weight parameter
        self.weight = nn.Parameter(torch.empty(
            (out_features, in_features),
            dtype=dtype
        ))

        # Alpha scaling parameter
        self.alpha = nn.Parameter(torch.ones(
            (1,),
            dtype=dtype
        ))
```

```python
        # Bias parameter
        if use_bias:
            self.bias = nn.Parameter(torch.empty(
                (out_features,),
                dtype=dtype
            ))
        else:
            self.register_parameter('bias', None)

        # Initialize parameters
        self.reset_parameters(kernel_init, bias_init, alpha_init)

    def reset_parameters(
        self,
        kernel_init: callable = None,
        bias_init: callable = None,
        alpha_init: callable = None
    ):
        """
        Initialize network parameters with specified or default initializers.
        """
        # Kernel (weight) initialization
        if kernel_init is None:
            kernel_init = nn.init.orthogonal_
        kernel_init(self.weight)

        # Bias initialization
        if self.use_bias:
            if bias_init is None:
                # Default: uniform initialization
                fan_in, _ = nn.init._calculate_fan_in_and_fan_out(self.weight)
                bound = 1 / math.sqrt(fan_in) if fan_in > 0 else 0
                nn.init.uniform_(self.bias, -bound, bound)
            else:
                bias_init(self.bias)

        # Alpha initialization (default to 1.0)
        if alpha_init is None:
            self.alpha.data.fill_(1.0)
        else:
            alpha_init(self.alpha)

    def forward(self, x: torch.Tensor) -> torch.Tensor:
        """
        Forward pass with squared Euclidean distance transformation.

        Args:
            x (torch.Tensor): Input tensor

        Returns:
            torch.Tensor: Transformed output
        """
```

```python
        # Ensure input and weight are in the same dtype
        x = x.to(self.dtype)

        # Compute dot product
        y = torch.matmul(x, self.weight.t())

        # Compute squared distances
        inputs_squared_sum = torch.sum(x**2, dim=-1, keepdim=True)
        kernel_squared_sum = torch.sum(self.weight**2, dim=-1)
        distances = inputs_squared_sum + kernel_squared_sum - 2 * y

        # Apply squared Euclidean distance transformation
        y = y ** 2 / (distances + self.epsilon)

        # Dynamic scaling
        scale = (math.sqrt(self.out_features) / math.log(1 + self.out_features)) ** self.alph
        y = y * scale

        # Add bias if used
        if self.use_bias:
            y += self.bias

        return y

    def extra_repr(self) -> str:
        """
        Extra representation of the module for print formatting.
        """
        return (f"in_features={self.in_features}, "
                f"out_features={self.out_features}, "
                f"bias={self.use_bias}")
```